\pdfoutput=1

\documentclass[runningheads]{llncs}
\usepackage{graphicx}
\usepackage{marvosym} 
\usepackage{xcolor}
\usepackage{enumerate}
\usepackage{xspace}
\usepackage{bm}
\usepackage{bbm}
\usepackage{amsfonts}
\usepackage[normalem]{ulem}
\usepackage{multirow}
\usepackage{amsmath}
\usepackage{enumitem}

\begin{document}


\title{A Self-guided Multimodal Approach to Enhancing Graph Representation Learning for Alzheimer's Diseases}

\titlerunning{Self-guided Multimodal GNN for AD}



\author{
Zhepeng Wang\inst{1}
\and
Runxue Bao\inst{2}
\and
Yawen Wu\inst{3}
\and
Guodong Liu\inst{4}
\and
Lei Yang\inst{1}
\and
Liang Zhan\inst{3}
\and
Feng Zheng\inst{5}
\and
Weiwen Jiang\inst{1}
\and
Yanfu Zhang\inst{6}
}
\authorrunning{Z. Wang et al.}

\institute{
\textsuperscript{1}George Mason University, \textsuperscript{2}GE Healthcare, \textsuperscript{3}University of Pittsburgh,\\
\textsuperscript{4}University of Maryland, \textsuperscript{5}Southern University of Science and Technology,\\ \textsuperscript{6}William and Mary
}

%
%

\maketitle   
\begin{abstract}
Graph neural networks (GNNs) are powerful machine learning models designed to handle irregularly structured data. However, their generic design often proves inadequate for analyzing brain connectomes in Alzheimer’s Disease (AD), highlighting the need to incorporate domain knowledge for optimal performance. Infusing AD-related knowledge into GNNs is a complicated task. Existing methods typically rely on collaboration between computer scientists and domain experts, which can be both time-intensive and resource-demanding. To address these limitations, this paper presents a novel self-guided, knowledge-infused multimodal GNN that autonomously incorporates domain knowledge into the model development process. Our approach conceptualizes domain knowledge as natural language and introduces a specialized multimodal GNN capable of leveraging this uncurated knowledge to guide the learning process of the GNN, such that it can improve the model performance and strengthen the interpretability of the predictions. To evaluate our framework, we curated a comprehensive dataset of recent peer-reviewed papers on AD and integrated it with multiple real-world AD datasets. Experimental results demonstrate the ability of our method to extract relevant domain knowledge, provide graph-based explanations for AD diagnosis, and improve the overall performance of the GNN. This approach provides a more scalable and efficient alternative to inject domain knowledge for AD compared with the manual design from the domain expert, advancing both prediction accuracy and interpretability in AD diagnosis.

\end{abstract}

\section{Introduction}
\label{sec:intro}
\begin{figure}[!t]
    \centering
    \includegraphics[width=\linewidth]{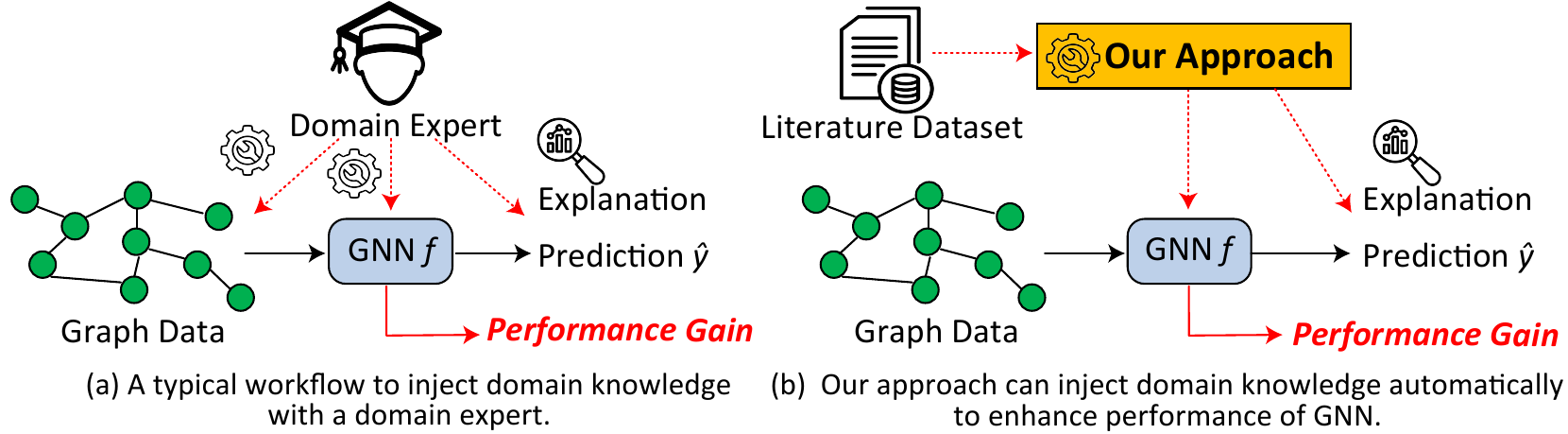}
    \caption{Illustration of designing AD-specific GNNs.}
    \label{fig: motivation}
\end{figure}

Graph neural networks (GNNs) have emerged as robust and versatile algorithms for diverse machine learning tasks involving graph or irregularly structured data, as evidenced by a plethora of studies~\cite{kipf2016semi,velivckovic2017graph,xu2018powerful,gilmer2017neural,xie2023accel,cui2022interpretable,peng2024maxk}. In recent years, there has been a discernible surge in research endeavors aimed at extending the applicability of GNNs towards addressing broader scientific and health-related challenges, such as drug prediction~\cite{xiong2021graph,lin2020kgnn}, precision medicine~\cite{wolterink60geometric,wang2022cell}, brain network analysis~\cite{tang2022contrastive,tang2023signed}, and protein structure prediction~\cite{xia2021geometric,tang2021commpool}. 

Despite the inherent capability of GNNs to capture crucial structural information within graphs, the direct application of generic GNNs to Alzheimer's Disease (AD) research is not straightforward. This is due to the distinctive properties of brain connectomes, such as fixed node numbers and orders of Regions of Interest (ROIs), which are not typically incorporated in standard GNN architectures. The pursuit of this research direction mandates a collaborative effort between neuroscientists and computer scientists to tailor GNNs specifically for AD, as shown in Fig.~\ref{fig: motivation}(a). However, the imperative for frequent interaction with domain experts contrasts with the overarching goal of minimizing manual intervention in machine learning algorithm development. The efficacy of the model may be constrained by the expertise and knowledge of the consulted experts, while the training of reliable domain experts incurs significant time and cost. Moreover, the scarcity of available experts poses further challenges, hindering model development efforts. Furthermore, the opaque nature of GNNs complicates the elucidation of decision-making processes, limiting the assistance that human experts can provide.

In an effort to reduce reliance on expert guidance in the development of GNNs for AD, our objective is to leverage self-collected, coarse, and uncurated domain knowledge instead of relying solely on meticulously curated expert knowledge. This approach raises several pertinent research questions: (1) How can uncurated domain knowledge, readily accessible through sources such as peer-reviewed publications on the internet, be effectively utilized by GNNs? (2) How can this knowledge be identified, curated, and tailored to suit specific datasets and GNN architectures? (3) How can such knowledge be effectively integrated into the model design to enhance performance and interoperability?

To address these challenges, we propose a self-guided multimodal approach to enhance graph representation learning for AD, as depicted in Fig.~\ref{fig: motivation}(b). Inspired by retrieval-augmented generation (RAG)~\cite{lewis2020retrieval,borgeaud2022improving,izacard2022few,luo2023augmented} techniques for large language models, we augment GNNs with extensive domain knowledge extracted from a self-built literature dataset, employing an end-to-end methodology to automatically discern graph-wise and knowledge-wise importance for guiding interpretable predictions. Initially, we integrate multimodal inputs (i.e., graph and natural language) and pretrain the multimodal GNN. Subsequently, we compute the importance scores of brain substructures and external knowledge, represented by explainable mask values. Finally, we refine the pretrained model through augmented brain connectomes guided by the masks to enhance performance.
Our method contributions can be summarized as follows:

\begin{itemize}[leftmargin=3mm]
    \item We propose an approach to automatically incorporate domain knowledge into GNNs to minimize human intervention, leveraging an external source of coarse and uncurated knowledge, and introducing a novel ``soft'' augmented retrieval generation method for integration with relevant subjects.
    \item Our approach aims to concurrently enhance the utility and interpretability of GNNs by identifying graph-wise and knowledge-wise explanations relevant to predictions. Moreover, the extracted explanations are tailored to the given dataset and GNNs thereby guiding graph augmentation for model fine-tuning.
    \item Experimental results on representative AD datasets demonstrate that our method can effectively inject domain knowledge into GNNs, thereby improving performance and facilitating the generation of customized explanations.
\end{itemize}

\section{Method}
\label{sec:method}

\begin{figure}[!t]
    \centering
    \includegraphics[width=\linewidth]{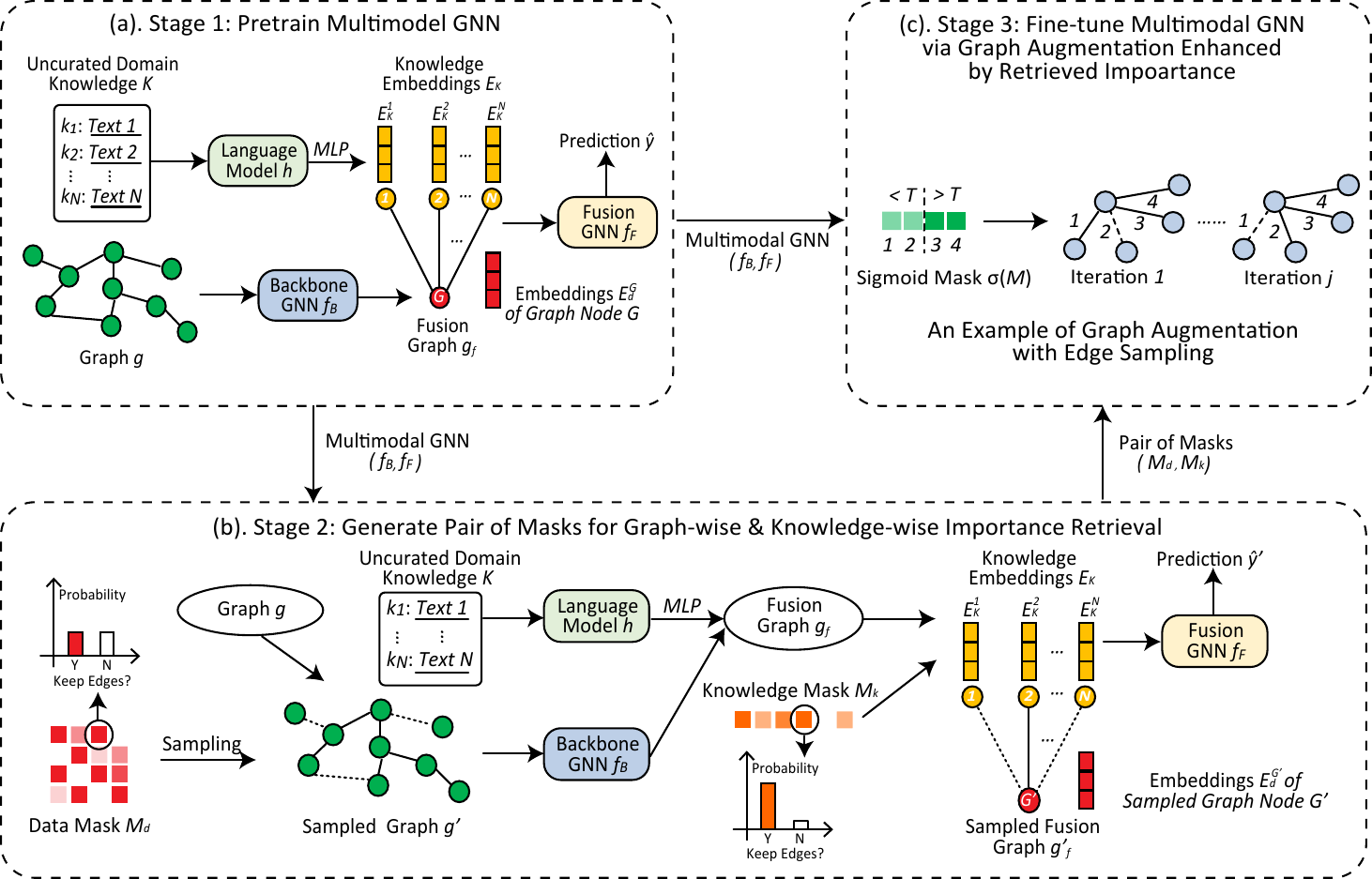}
    \caption{Design overview of our approach.}
    \label{fig: MM_GNN}
\end{figure}

\noindent\textbf{Overview.}
\label{subsec:problem form}
To integrate external knowledge into GNNs, we will fuse the embeddings from graphs and external knowledge guided by their relevance to the prediction. More specifically, we learn a pair of masks to characterize the relevance and use the masked embeddings to fine-tune the prediction model. We first formulate a multimodal GNN denoted as $f$, which accepts brain connectome data $g$ alongside natural language data $\mathcal{K}$ as inputs. In essence, this entails augmenting canonical GNNs with external knowledge data. Subsequently, we undertake the pretraining of $f$ using gathered data. We then identify significant substructures within brain connectomes and pertinent knowledge relevant to prediction tasks. These are delineated by real-valued masks parameterized by $\bm{\alpha}$ and $\bm{\beta}$. By computing the masks jointly, we facilitate a ``soft'' retrieval mechanism from the knowledge data, wherein the masks function as explanations for both graph-wise and knowledge-wise aspects of predictions made by $f$. Ultimately, we refine the parameters of $f$ through a process of fine-tuning, leveraging graph augmentation in conjunction with the ``soft'' retrieval outcomes. Specifically, we utilize the computed masks to guide edge sampling, thereby preserving critical information while augmenting data diversity. The design overview is shown in Fig.~\ref{fig: MM_GNN} and further elucidation of our approach follows.

\noindent\textbf{Multimodal GNN equipped with external knowledge.}
We propose the integration of canonical GNNs with external knowledge, culminating in a multimodal GNN. Specifically, our multimodal GNN comprises a backbone GNN denoted as $f_{B}$ and a fusion GNN denoted as $f_{F}$. Given the graph data $g$, the backbone GNN $f_{B}$ is employed to generate a node embedding $E_{d}^{G} = f_{B}(g)$. 
The uncurated domain knowledge $\mathcal{K}$ is represented as a collection of language sequences, denoted as $\mathcal{K} = \{k_1, k_2, ..., k_{N}\}$, where each knowledge item $k_i$ corresponds to a summarization of domain knowledge. For instance, the abstract of a paper focusing on Alzheimer's Disease (AD) can be considered as a language sequence $k_i$ within the uncurated domain knowledge $\mathcal{K}$ for AD. To integrate $\mathcal{K}$ with $E_{d}^{G}$, we compute $E_{\mathcal{K}}^{i} = MLP(h(k_{i})), \forall k_{i} \in \mathcal{K}$, where a pretrained language model $h$ followed by a multi-layer perceptron $MLP$ is utilized to generate the language embedding. We denote the knowledge embeddings collectively as $E_{\mathcal{K}} = \{E_{\mathcal{K}}^{i}| i = idx(k_i), k_i \in \mathcal{K}\}$ with $|\mathcal{K}| = N$.
Utilizing $E_{d}^{G}$ and $E_{\mathcal{K}}$, a fusion graph $g_f$ is constructed by connecting node $G$ with feature $E_{d}^{G}$ to each node $i$ in $\mathcal{K}$ with feature $E_{\mathcal{K}}^{i}$. This allows us to modify cross-modal edges to represent retrieval. The final prediction $\hat{y}$ for the target task is generated by the fusion GNN $f_{F}$, where $\hat{y} = f_{F}(g_f)$. Subsequently, following the forward propagation of the multimodal GNN $f$ as defined, the model can be trained end-to-end using multimodal inputs $g$ and $\mathcal{K}$, leveraging any arbitrary loss function $\mathcal{L}$ specific to the target task. The design of our multimodal GNN is illustrated in Fig.~\ref{fig: MM_GNN}(a).

\noindent\textbf{Graph-wise and knowledge-wise importance retrieval using masks.}
Given the pretrained GNN, we aim to identify the importance score represented by real-value masks of substructures in the graphs and the retrieved knowledge, to indicate the crucial components contributing to the prediction, and the details are shown in Fig.~\ref{fig: MM_GNN}(b).
Following the fashion in previous works, we define the explanation of predictions on the multimodal input $g$ and $\mathcal{K}$ as a subgraph $\tilde{g}$ and a subset $\tilde{\mathcal{K}}$. Since directly sampling them from $g$ and $\mathcal{K}$ is computationally prohibitive and indifferentiable, we parameterized the sampling process as two types of learnable masks combined with Gumbel Softmax function~\cite{jang2016categorical}, i.e., data mask $M_{d}$ parameterized with $\bm{\alpha}$ on the graph $g$ and knowledge mask $M_{k}$ parameterized with $\bm{\beta}$ on the fusion graph $g_f$. More specifically, for a given mask $M$ ($M_{d}$ or $M_{k}$) parameterized with $\bm{\phi}$ ($\bm{\alpha}$ for $M_{d}$ or $\bm{\beta}$ for $M_{k}$), Gumbel Softmax function is applied to arbitrary entry within $M$ as $m_{i,j} = \frac{\exp{[(\phi_{i,j} + g_{i,j})/\tau]}}{\exp{[(\phi_{i,j} + g_{i,j})]} + \exp{(g_{i,j}^{\prime})}}$, where $\phi_{i,j}$ denotes the entry value indexed with $(i,j)$ in $M$. $g_{i,j}$ and $g_{i,j}^{\prime}$ are two independent random noise sampled from Gumbel distribution $Gumbel(0,1)$ controled by temperature $\tau$.

Given graph $g = (V_g, \bm{W}_g)$, where $V_g$ denotes the node set of $g$ and $\bm{W}_g \in \mathbb{R}^{|V_g| \times |V_g|}$ represents the weighted adjacency matrix of $g$, the matrix $M_{d}^{\prime}$ sampled from $M_{d}$ is applied to $\bm{W}_g$ to approximate the sampling of edges as $ \bm{W}_g^{\prime} = M_{d}^{\prime} \odot \bm{W}_g$, where $\odot$ is the element-wise matrix multiplication. With $\bm{W}_g^{\prime}$, we sample $g^{\prime} = (V_g, \bm{W}_g^{\prime})$, and calculate the node embedding as $E_{d}^{G^{\prime}}=f_{B}(g^{\prime})$. Node embedding $E_{d}^{G^{\prime}}$ is then used together with the knowledge embedding $E_{\mathcal{K}}$ to construct the fusion graph $g_f = (V_f, \bm{A}_f)$, where $V_f$ denotes the node set of $g_f$ and $\bm{A}_f \in \mathbb{R}^{|V_f|-1 }$ represents the unweighted adjacency matrix of $g_f$, where all the entries are one.
Similarly, for fusion graph $g_f$, we apply the sampled mask $M_{k}^{\prime}$ to $\bm{A}_f$ to approximate the sampling of edges as $\bm{A}_{f}^{\prime} = M_{k}^{\prime} \odot \bm{A}_{f}$ and get the sampled fusion graph $g_{f}^{\prime}$. The final prediction is calculated by $f_{F}(g_{f}^{\prime})$.

We define $\mathcal{L}_{exp}$ to learn the masks with hyperparameters $\lambda_1, \lambda_2, \lambda_3, \lambda_4$,
\begin{equation}\label{eq:mask_loss}
    \mathcal{L}_{exp} = \lambda_1 \mathcal{L}_{mask} + \lambda_2 \mathcal{L}_{clf} + \lambda_3 \mathcal{L}_{spas} + \lambda_4 \mathcal{L}_{disc}.
\end{equation}
$\mathcal{L}_{mask} = - \sum_{c=1}^{C}\mathbbm{1}[\hat{y}=c]\log P_{f}(\hat{y}^{\prime}=\hat{y}|g^{\prime}, g^{\prime}_{f})$ denotes the disagreement between the prediction $\hat{y}$ made on the original inputs and the prediction $\hat{y}^{\prime}$ made on the sampled inputs. $C$ is the class number in the target task. 
$\mathcal{L}_{clf} =  - \sum_{c=1}^{C}\mathbbm{1}[y=c]\log P_{f}(\hat{y}^{\prime}=y|g^{\prime}, g^{\prime}_{f})$ encourages the consistency between $\hat{y}^{\prime}$ and the label $y$.
$\mathcal{L}_{spas} = \frac{1}{|M_{d}|} \sum_{i,j} \sigma (M_{d}^{i, j}) + \frac{1}{|M_{k}|} \sum_{i,j} \sigma (M_{k}^{i, j})$ controls the sparsity of the two masks, where $\sigma(\cdot)$ represents the sigmoid function. $|M_{d}|$ and $|M_{k}|$ are the number of entries in $M_{d}$ and $M_{k}$, respectively. 
$\mathcal{L}_{disc} = \sum_{i,j} \frac{\mathcal{L}_{ent} (\sigma(M_{d}^{i, j}))}{|M_{d}|} + \sum_{i,j} \frac{\mathcal{L}_{ent} (\sigma(M_{k}^{i, j}))}{|M_{k}|}$ promotes the discreteness of the two masks, where $\mathcal{L}_{ent} (\cdot)$ is the binary entropy function.

This section only considers one pair of $M_{d}$ and $M_{k}$ for simplicity. However, we can generalize it to multiple pairs cases. For instance, if genders are provided, we can define $M^{female} = \{M_{d}^{female}, M_{k}^{female}\}$ and $M^{male} = \{M_{d}^{male}, M_{k}^{male}\}$, and train each pair of masks separately.

\noindent\textbf{Graph augmentation enhanced by retrieved importance.}
We propose a graph augmentation method guided by the crucial components in prediction, i.e., the learned masks, to fine-tune the pretrained multimodal GNN $f$, aiming to enhance model performance. Specifically, during fine-tuning, we employ a threshold $T$ on $\sigma(M)$ to guide edge sampling as follows: for any arbitrary entry value $m_{i}$, we retain the corresponding edge when $m_{i} \geq T$; otherwise, we randomly sample the corresponding edge with a probability of $0.5$ for each iteration. A schematic illustration of this process is provided in Fig.~\ref{fig: MM_GNN}(c), where $m_1$ and $m_2$ are both less than $T$, while $m_3$ and $m_4$ exceed $T$. Consequently, edges 3 and 4 are retained across both iteration 1 and $j$, whereas edge 1 exists solely in iteration 1, and edge 2 appears only in iteration $j$. This methodology can be uniformly applied to all computed masks, thereby substantially increasing input diversity to $f$ concerning less significant edges, while retaining crucial edges for prediction, as indicated by the entry values within the masks.

\section{Experiments}
\label{sec:experiments}
\noindent\textbf{Datasets and Settings.}
We evaluate our approach on two AD datasets, i.e., OASIS~\cite{lamontagne2019oasis} and ADNI-D~\cite{mackin2021late}. For each, we include two graph datasets in different modalities, where one is derived from DTI imaging while the other one is derived from fMRI imaging.

\begin{itemize}[leftmargin=3mm,label=*]
    \item \textbf{\underline{OASIS}} contains 815 subjects. 155 subjects are diagnosed with AD and the others are seronegative controls. For each subject, there are 132 regions of interest (ROIs) based on Harvard-Oxford Atlas~\cite{desikan2006automated} and AAL Atlas~\cite{tzourio2002automated}. There are 459 females, including 66 AD patients and 393 seronegative controls, and 356 males, including 89 AD patients and 267 seronegative controls.
    \item \textbf{\underline{ADNI-D}} contains 340 subjects. 154 patients are diagnosed with mild cognitive impairment (MCI), which is the early stage of AD. The others are seronegative controls. For each subject, 85 ROIs are derived from T1-weighted MRI using FreeSurfer (V6.5)~\cite{dale1999cortical}. There are 210 females, including 84 MCI patients and 126 seronegative controls, and 130 males, including 70 MCI patients and 60 seronegative controls.
\end{itemize}

For domain knowledge $\mathcal{K}$ of AD, we collected 20,108 records related to AD in the last 20 years from PubMed~\footnote{https://pubmed.ncbi.nlm.nih.gov/}. Each record consists of the title and the abstract of a paper. We use BERT-Large~\cite{devlin2018bert} to encode $\mathcal{K}$. BERT-Large is pretrained with whole word masking on the uncased version of the training corpus.

We evaluate GCN~\cite{kipf2016semi}, GINE~\cite{xu2018powerful}, and GAT~\cite{velivckovic2017graph} as the architecture of both the backbone and fusion GNN within the multimodel GNN (MM-GNN). Given the prior knowledge about the gender distribution of the dataset, we introduce two pairs of masks for graph-wise and knowledge-wise importance retrieval, where one is for males and the other is for females. Please refer to the supplementary materials for training details. 



\noindent\textbf{Main Results}. The proposed MM-GNN significantly outperforms the plain version of GNN, and the MM-GNN with fine-tuning further improves the performance.
Table~\ref{tab:Table_Main} reports the main results for the evaluation of our method. 
\begin{table*}[!t]
\centering
\footnotesize
\caption{Main results.}
\label{tab:Table_Main}
\resizebox{\textwidth}{!}{\begin{tabular}{cllllllllllll}
\hline
\multirow{2}{*}{Method} & \multicolumn{3}{c}{OASIS (DTI)} & \multicolumn{3}{c}{OASIS (fMRI)} & \multicolumn{3}{c}{ADNI-D (DTI)} & \multicolumn{3}{c}{ADNI-D (fMRI)} \\ \cline{2-13} 
 & \multicolumn{1}{c}{ACC} & \multicolumn{1}{c}{AUC} & \multicolumn{1}{c}{F1} & \multicolumn{1}{c}{ACC} & \multicolumn{1}{c}{AUC} & \multicolumn{1}{c}{F1} & \multicolumn{1}{c}{ACC} & \multicolumn{1}{c}{AUC} & \multicolumn{1}{c}{F1} & \multicolumn{1}{c}{ACC} & \multicolumn{1}{c}{AUC} & \multicolumn{1}{c}{F1} \\ \hline
GCN & 0.6466 & 0.5246 & 0.4840 & 0.6245 & 0.5376 & 0.5003 & 0.5765 & 0.4642 & 0.4619 & 0.5618 & 0.4978 & 0.5301 \\
MM-GCN (ours) & 0.7141 & 0.5220 & 0.5639 & 0.7387 & 0.5774 & 0.5962 & 0.6294 & 0.5629 & 0.6133 & 0.6235 & 0.5852 & 0.6111 \\
MM-GCN-F (ours) & \textbf{0.7362} & \textbf{0.5381} & \textbf{0.5805} & \textbf{0.7436} & \textbf{0.5774} & \textbf{0.5989} & \textbf{0.6471} & \textbf{0.5821} & \textbf{0.6406} & \textbf{0.6441} & \textbf{0.5936} & \textbf{0.6367} \\ \hline
GINE & 0.7448 & 0.5238 & 0.5332 & 0.6405 & 0.5476 & 0.5251 & 0.5765 & 0.5966 & 0.5242 & 0.5676 & 0.5862 & 0.5439 \\
MM-GINE (ours) & 0.7521 & \textbf{0.5777} & 0.6004 & \textbf{0.7558} & 0.5641 & 0.5939 & 0.6706 & 0.6286 & 0.6610 & \textbf{0.6765} & 0.6450 & \textbf{0.6671} \\
MM-GINE-F (ours) & \textbf{0.7926} & 0.5733 & \textbf{0.6245} & 0.7534 & \textbf{0.6030} & \textbf{0.6121} & \textbf{0.6853} & \textbf{0.6720} & \textbf{0.6774} & 0.6647 & \textbf{0.6716} & 0.6556 \\ \hline
GAT & 0.5975 & 0.5340 & 0.4925 & 0.6847 & 0.4809 & 0.5184 & 0.5529 & 0.5300 & 0.4617 & 0.5824 & 0.4994 & 0.4512 \\
MM-GAT (ours) & 0.7436 & \textbf{0.6098} & 0.5983 & 0.7632 & 0.5828 & 0.6040 & 0.6588 & 0.6164 & 0.6354 & 0.6559 & 0.6072 & 0.6252 \\
MM-GAT-F  (ours) & \textbf{0.7460} & 0.6085 & \textbf{0.6022} & \textbf{0.7779} & \textbf{0.6008} & \textbf{0.6143} & \textbf{0.6706} & \textbf{0.6204} & \textbf{0.6557} & \textbf{0.6618} & \textbf{0.6102} & \textbf{0.6446} \\ \hline
\end{tabular}
}

\end{table*}

As shown in Table~\ref{tab:Table_Main}, the performance can be improved significantly when the domain knowledge is injected into the inference of GNN. For example, on the two datasets of ADNI-D, compared with vanilla GCN, 10.08\%, 19.41\% and 24.03\% relative improvement is gained in ACC, AUC and F1 on average, respectively. 
Moreover, the performance can be further enhanced when fine-tuning the multimodal GNN guided by the generated masks. For instance, on the two datasets of ADNI-D, compared with plain MM-GCN, 3.06\%, 2.43\% and 4.32\% relative improvement can be achieved in ACC, AUC and F1 on average, respectively.


\label{subsec:ablation}
\begin{figure*}[!t]
    \centering
    \includegraphics[width=\linewidth]{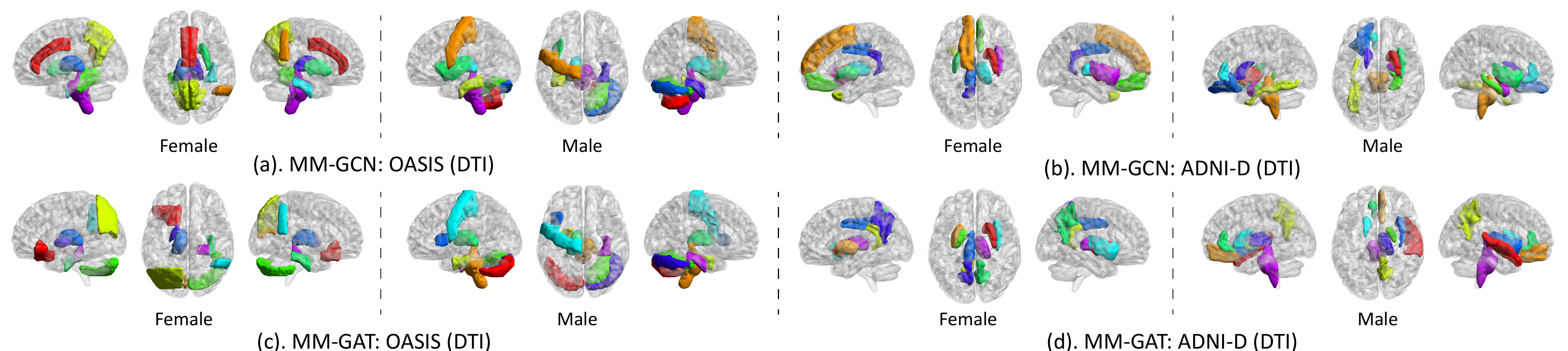}
    \caption{Brain saliency maps identified by our approach. Top 10 salient ROIs are highlighted.}
    \label{fig: salient_roi}
\end{figure*}
\noindent\textbf{Qualitative Analysis on Graph-wise Masks}. To analyze the generated graph-wise masks, we calculate the importance score of each ROI based on the graph-wise mask and highlight the top 10 salient ROIs on the brain saliency maps in Fig.~\ref{fig: salient_roi}. Note that the color in each brain saliency map of Fig.~\ref{fig: salient_roi} is applied to distinguish different ROIs only.



As shown in Fig.~\ref{fig: salient_roi}, it is clear that the salient ROIs are distinct for males and females, which is consistent with the existing studies~\cite{subramaniapillai2021sex,williamson2022sex}. Moreover, MM-GNNs with different architectures can also show a distinct preference for their prediction even if they are trained on the same dataset. For instance, MM-GCN and MM-GAT share 5 salient ROIs and have 5 unique salient ROIs. Such kinds of insights are usually difficult for human experts to extract but can be automatically captured by our approach.



\begin{figure*}[!t]
    \centering
    \includegraphics[width=\linewidth]{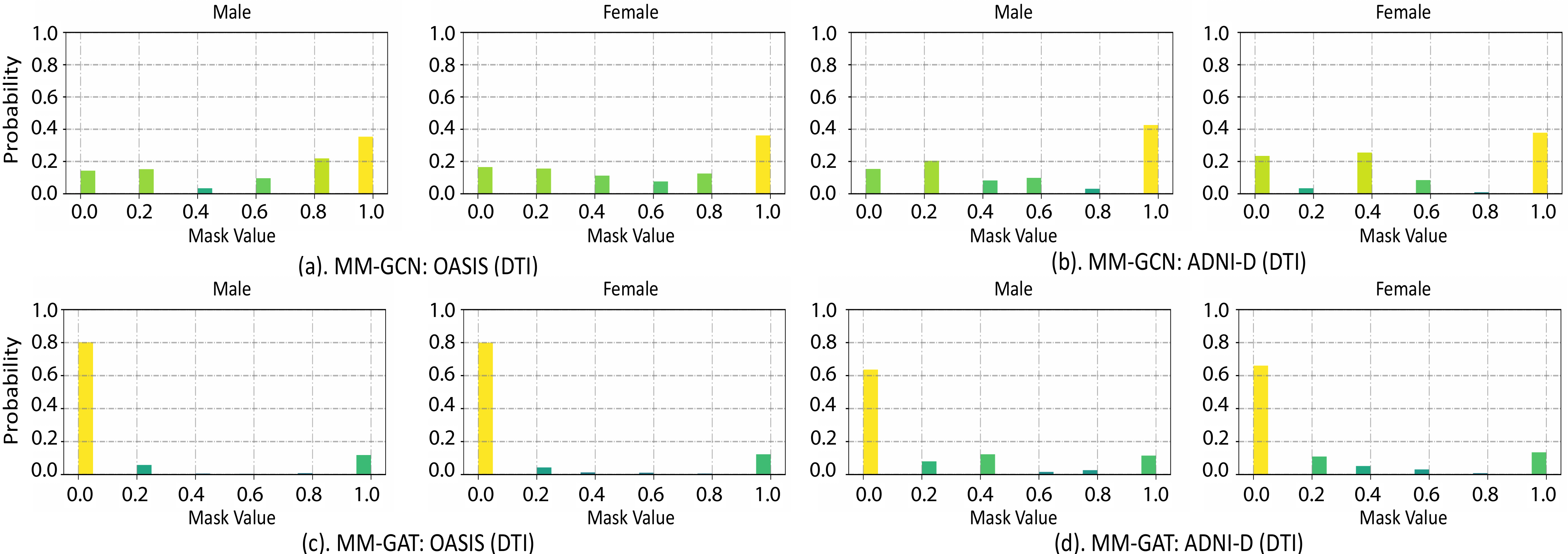}
    \caption{Distribution of importance scores of domain knowledge from our approach.}
    \label{fig: salient_knowledge}
\end{figure*}
\noindent\textbf{Qualitative Analysis on Knowledge-wise Masks}. Our approach can automatically adapt its way to employ the domain knowledge to the change in the gender, backbone and dataset. Fig.~\ref{fig: salient_knowledge} shows the distribution of values within the knowledge-wise masks, which can be interpreted as the distribution of importance scores of domain knowledge. And we have several observations. First, males and females have different ways to leverage the given domain knowledge for inference. Second, different architectures of MM-GNN show different patterns on the knowledge-wise mask. Compared with MM-GCN, MM-GAT tends to have a more sparse distribution. Last, even when the domain is the same (i.e., AD), our approach can propose different paradigms to leverage the domain knowledge when the dataset is different.

\noindent\textbf{Ablation Study on the Size of Domain Knowledge}. We conducted the ablation study on the size of the domain knowledge, where we randomly sampled 1\% and 10\% records from the complete set of domain knowledge $\mathcal{K}$ to train the MM-GNN. The results are shown in Table~\ref{tab:Table_NLP_Size} and we have two observations. First, the performance improvement is minor and even suffers from degradation when increasing the size of domain knowledge from about 200 (1\%) to about 2,000 (10\%). Second, the benefits from a larger set of domain knowledge become obvious when increasing the size of domain knowledge to more than 20,000 (100\%). Therefore, we can conclude that although the performance may have little improvement when increasing the set of domain knowledge to a moderate size, a noticeable boost in performance can be observed when increasing the size of domain knowledge to a much larger size.

\begin{table*}[!t]
\centering
\tabcolsep=0.3em
\caption{Ablation study on the size of domain knowledge.}
\label{tab:Table_NLP_Size}
\resizebox{\textwidth}{!}{\begin{tabular}{cllllllllllll}
\hline
\multirow{2}{*}{Method} & \multicolumn{3}{c}{OASIS (DTI)} & \multicolumn{3}{c}{OASIS (fMRI)} & \multicolumn{3}{c}{ADNI-D (DTI)} & \multicolumn{3}{c}{ADNI-D (fMRI)} \\ \cline{2-13} 
 & \multicolumn{1}{c}{ACC} & \multicolumn{1}{c}{AUC} & \multicolumn{1}{c}{F1} & \multicolumn{1}{c}{ACC} & \multicolumn{1}{c}{AUC} & \multicolumn{1}{c}{F1} & \multicolumn{1}{c}{ACC} & \multicolumn{1}{c}{AUC} & \multicolumn{1}{c}{F1} & \multicolumn{1}{c}{ACC} & \multicolumn{1}{c}{AUC} & \multicolumn{1}{c}{F1} \\ \hline
MM-GCN (100\%) & \textbf{0.7141} & 0.5220 & \textbf{0.5639} & \textbf{0.7387} & \textbf{0.5774} & \textbf{0.5962} & \textbf{0.6294} & \textbf{0.5629} & \textbf{0.6133} & \textbf{0.6235} & \textbf{0.5852} & \textbf{0.6111} \\
MM-GCN (10\%) & 0.6822 & 0.5367 & 0.4852 & 0.6405 & 0.5636 & 0.4921 & 0.5941 & 0.5191 & 0.5833 & 0.6118 & 0.5682 & 0.5656 \\
MM-GCN (1\%) & 0.6613 & \textbf{0.5388} & 0.4993 & 0.6429 & 0.5573 & 0.5264 & 0.5912 & 0.5277 & 0.5871 & 0.6235 & 0.5610 & 0.5755 \\ \hline
MM-GINE (100\%) & 0.7521 & \textbf{0.5777} & \textbf{0.6004} & 0.7558 & \textbf{0.5641} & \textbf{0.5939} & 0.6706 & \textbf{0.6286} & \textbf{0.6610} & \textbf{0.6765} & 0.6450 & \textbf{0.6671} \\
MM-GINE (10\%) & 0.8184 & 0.5546 & 0.5115 & 0.7644 & 0.5294 & 0.5801 & 0.6676 & 0.6206 & 0.6479 & 0.6059 & \textbf{0.6806} & 0.5754 \\
MM-GINE (1\%) & \textbf{0.8221} & 0.5206 & 0.5152 & \textbf{0.8221} & 0.5078 & 0.5244 & \textbf{0.6735} & 0.6184 & 0.6608 & 0.6206 & 0.6937 & 0.5933 \\ \hline
MM-GAT (100\%) & 0.7436 & \textbf{0.6098} & \textbf{0.5983} & 0.7632 & \textbf{0.5828} & \textbf{0.6040} & \textbf{0.6588} & \textbf{0.6164} & \textbf{0.6354} & \textbf{0.6559} & \textbf{0.6072} & \textbf{0.6252} \\
MM-GAT (10\%) & 0.7485 & 0.5665 & 0.5896 & 0.8184 & 0.5317 & 0.5012 & 0.6088 & 0.5578 & 0.5824 & 0.5559 & 0.4828 & 0.5475 \\
MM-GAT (1\%) & \textbf{0.7718} & 0.5562 & 0.5893 & \textbf{0.8221} & 0.5176 & 0.5290 & 0.6324 & 0.5822 & 0.5860 & 0.6118 & 0.5470 & 0.5306 \\ \hline
\end{tabular}
}
\end{table*}



%
\section{Related Work}
\label{sec:related_work}

\noindent\textbf{Graph Neural Network (GNN)}. Deep neural networks (DNNs) have achieved great success in processing data with regular formats, such as the convolutional neural networks for image~\cite{wu2023synthetic,peng2023autorep,wu2021decentralized,zhang2022toward,zhang2024neurodegenerative,wu2021enabling,wu2024auto} and recurrent neural networks for sequence data~\cite{graves2012long,greff2016lstm,schmidhuber2007training}. However, the irregular structures of graph data pose unique challenges to applying DNNs to handling graph data. Therefore, a great amount of effort has been made to the research to design the graph neural networks, whose core is the iterative update of node representations by integrating the features of neighboring nodes~\cite{bruna2013spectral,kipf2016variational,bronstein2017geometric,hamilton2017inductive,gilmer2017neural,battaglia2018relational}. Graph convolutional networks (GCNs)~\cite{kipf2016semi}, the most popular GNN architecture, are recognized for their simplicity and effectiveness, employing a layer-wise linear transformation matrix in node representation refinement.
Graph attention networks (GAT)~\cite{velivckovic2017graph} marks a significant evolution in GNNs, addressing GCNs' limitations by employing an attention mechanism~\cite{bahdanau2014neural,vaswani2017attention} to evaluate neighbor significance, which is widely used in applications of natural language processing~\cite{jin2024learning} and computer vision~\cite{jin2023visual,zhang2024brain,wu2020intermittent,wu2020enabling,wu2021federated,wang2021lightweight,wu2021federated,wu2022distributed}. Graph isomorphism Network (GIN)~\cite{xu2018powerful} is another critical development in GNN research, focusing on the implementation of permutation invariance to ensure output consistency irrespective of node index order. These breakthroughs in the design of GNN have made it one of the most popular algorithms to handle graph data. GNNs have now been widely used in diverse graph applications such as community detection~\cite{shchur2019overlapping,luo2021detecting,wu2022clare}, disease diagnosis~\cite{tang2023comprehensive} and molecule property prediction~\cite{fouss2007random}.


\noindent\textbf{Understanding Graph Data with Large Language Models (LLMs)}. LLMs have demonstrated significant capability across a variety of tasks, including classification~\cite{gao2024transfer,zhang2024pruning}, question answering~\cite{wang2024infuserki,yu2024dynamic}, code generation~\cite{wang2024unlocking}, and named entity recognition~\cite{zhao2023survey}, etc. However, their proficiency in handling irregularly structured data, especially graph data, is an emerging yet underdeveloped area. The main challenges of applying LLMs to graph data are the lack of a solid method to describe the graph data in natural language without information loss and the effective prompt techniques to elicit the pretrained LLM to generate appropriate responses for graph-related applications. Recent studies~\cite{guo2023gpt4graph} have investigated the integration of LLMs with graph data to improve their effectiveness in graph mining tasks. The results show that while LLMs have some ability to handle graph-structured data, they still fall short of specialized graph-oriented models, indicating the need for further development. Moreover, there also emerges works to extend the usage of LLMs to graph tasks like selecting graph processors~\cite{zhang2023graph}, generating state-of-the-art (SOTA) graph embeddings~\cite{ye2023natural}, and creating prompts for graph inputs~\cite{jiang2023structgpt,han2023pive,qian2023can}. Another relevant application is representation learning on text-attributed graphs (TAGs)~\cite{he2023explanations}, which customized the prompts to ask the LLM to generate both predictions and text explanations for each node.

\section{Conclusion}
We introduce a self-guided approach to autonomously integrate domain knowledge into GNNs to harness collected uncurated AD knowledge. 
Our approach can effectively extract curated knowledge and explanations on graphs for AD and guide the fine-tuning to improve the performance of GNNs. Extensive experiments on real-world AD datasets demonstrate the effectiveness of our method.
\label{sec:conclusion}

\bibliographystyle{splncs04}
\bibliography{ref, arxiv}

\begin{thebibliography}{10}
\providecommand{\url}[1]{\texttt{#1}}
\providecommand{\urlprefix}{URL }
\providecommand{\doi}[1]{https://doi.org/#1}

\bibitem{bahdanau2014neural}
Bahdanau, D., Cho, K., Bengio, Y.: Neural machine translation by jointly learning to align and translate. arXiv preprint arXiv:1409.0473  (2014)

\bibitem{battaglia2018relational}
Battaglia, P.W., Hamrick, J.B., Bapst, V., Sanchez-Gonzalez, A., Zambaldi, V., Malinowski, M., Tacchetti, A., Raposo, D., Santoro, A., Faulkner, R., et~al.: Relational inductive biases, deep learning, and graph networks. arXiv preprint arXiv:1806.01261  (2018)

\bibitem{borgeaud2022improving}
Borgeaud, S., Mensch, A., Hoffmann, J., Cai, T., Rutherford, E., Millican, K., Van Den~Driessche, G.B., Lespiau, J.B., Damoc, B., Clark, A., et~al.: Improving language models by retrieving from trillions of tokens. In: International conference on machine learning. pp. 2206--2240. PMLR (2022)

\bibitem{bronstein2017geometric}
Bronstein, M.M., Bruna, J., LeCun, Y., Szlam, A., Vandergheynst, P.: Geometric deep learning: going beyond euclidean data. IEEE Signal Processing Magazine  \textbf{34}(4),  18--42 (2017)

\bibitem{bruna2013spectral}
Bruna, J., Zaremba, W., Szlam, A., LeCun, Y.: Spectral networks and locally connected networks on graphs. arXiv preprint arXiv:1312.6203  (2013)

\bibitem{cui2022interpretable}
Cui, H., Dai, W., Zhu, Y., Li, X., He, L., Yang, C.: Interpretable graph neural networks for connectome-based brain disorder analysis. In: International Conference on Medical Image Computing and Computer-Assisted Intervention. pp. 375--385. Springer (2022)

\bibitem{dale1999cortical}
Dale, A.M., Fischl, B., Sereno, M.I.: Cortical surface-based analysis: I. segmentation and surface reconstruction. Neuroimage  \textbf{9}(2),  179--194 (1999)

\bibitem{desikan2006automated}
Desikan, R.S., S{\'e}gonne, F., Fischl, B., Quinn, B.T., Dickerson, B.C., Blacker, D., Buckner, R.L., Dale, A.M., Maguire, R.P., Hyman, B.T., et~al.: An automated labeling system for subdividing the human cerebral cortex on mri scans into gyral based regions of interest. Neuroimage  \textbf{31}(3),  968--980 (2006)

\bibitem{devlin2018bert}
Devlin, J., Chang, M.W., Lee, K., Toutanova, K.: Bert: Pre-training of deep bidirectional transformers for language understanding. arXiv preprint arXiv:1810.04805  (2018)

\bibitem{fouss2007random}
Fouss, F., Pirotte, A., Renders, J.M., Saerens, M.: Random-walk computation of similarities between nodes of a graph with application to collaborative recommendation. IEEE Transactions on knowledge and data engineering  \textbf{19}(3),  355--369 (2007)

\bibitem{gao2024transfer}
Gao, Y., Bao, R., Ji, Y., Sun, Y., Song, C., Ferraro, J.P., Ye, Y.: Transfer learning with clinical concept embeddings from large language models. arXiv preprint arXiv:2409.13893  (2024)

\bibitem{gilmer2017neural}
Gilmer, J., Schoenholz, S.S., et~al.: Neural message passing for quantum chemistry. In: International Conference on Machine Learning. pp. 1263--1272. PMLR (2017)

\bibitem{graves2012long}
Graves, A., Graves, A.: Long short-term memory. Supervised sequence labelling with recurrent neural networks pp. 37--45 (2012)

\bibitem{greff2016lstm}
Greff, K., Srivastava, R.K., Koutn{\'\i}k, J., Steunebrink, B.R., Schmidhuber, J.: Lstm: A search space odyssey. IEEE transactions on neural networks and learning systems  \textbf{28}(10),  2222--2232 (2016)

\bibitem{guo2023gpt4graph}
Guo, J., Du, L., Liu, H.: Gpt4graph: Can large language models understand graph structured data? an empirical evaluation and benchmarking. arXiv preprint arXiv:2305.15066  (2023)

\bibitem{hamilton2017inductive}
Hamilton, W.L., Ying, R., Leskovec, J.: Inductive representation learning on large graphs. arXiv preprint arXiv:1706.02216  (2017)

\bibitem{han2023pive}
Han, J., Collier, N., Buntine, W., Shareghi, E.: Pive: Prompting with iterative verification improving graph-based generative capability of llms. arXiv preprint arXiv:2305.12392  (2023)

\bibitem{he2023explanations}
He, X., Bresson, X., Laurent, T., Hooi, B.: Explanations as features: Llm-based features for text-attributed graphs. arXiv preprint arXiv:2305.19523  (2023)

\bibitem{izacard2022few}
Izacard, G., Lewis, P., Lomeli, M., Hosseini, L., Petroni, F., Schick, T., Dwivedi-Yu, J., Joulin, A., Riedel, S., Grave, E.: Few-shot learning with retrieval augmented language models. arXiv preprint arXiv:2208.03299  (2022)

\bibitem{jang2016categorical}
Jang, E., Gu, S., Poole, B.: Categorical reparameterization with gumbel-softmax. arXiv preprint arXiv:1611.01144  (2016)

\bibitem{jiang2023structgpt}
Jiang, J., Zhou, K., Dong, Z., Ye, K., Zhao, W.X., Wen, J.R.: Structgpt: A general framework for large language model to reason over structured data. arXiv preprint arXiv:2305.09645  (2023)

\bibitem{jin2024learning}
Jin, C., Che, T., Peng, H., Li, Y., Pavone, M.: Learning from teaching regularization: Generalizable correlations should be easy to imitate. arXiv preprint arXiv:2402.02769  (2024)

\bibitem{jin2023visual}
Jin, C., Huang, T., Zhang, Y., Pechenizkiy, M., Liu, S., Liu, S., Chen, T.: Visual prompting upgrades neural network sparsification: A data-model perspective. arXiv preprint arXiv:2312.01397  (2023)

\bibitem{kipf2016semi}
Kipf, T.N., Welling, M.: Semi-supervised classification with graph convolutional networks. arXiv preprint arXiv:1609.02907  (2016)

\bibitem{kipf2016variational}
Kipf, T.N., Welling, M.: Variational graph auto-encoders. arXiv preprint arXiv:1611.07308  (2016)

\bibitem{lamontagne2019oasis}
LaMontagne, P.J., Benzinger, T.L., Morris, J.C., Keefe, S., Hornbeck, R., Xiong, C., Grant, E., Hassenstab, J., Moulder, K., Vlassenko, A.G., et~al.: Oasis-3: longitudinal neuroimaging, clinical, and cognitive dataset for normal aging and alzheimer disease. MedRxiv pp. 2019--12 (2019)

\bibitem{lewis2020retrieval}
Lewis, P., Perez, E., Piktus, A., Petroni, F., Karpukhin, V., Goyal, N., K{\"u}ttler, H., Lewis, M., Yih, W.t., Rockt{\"a}schel, T., et~al.: Retrieval-augmented generation for knowledge-intensive nlp tasks. Advances in Neural Information Processing Systems  \textbf{33},  9459--9474 (2020)

\bibitem{lin2020kgnn}
Lin, X., Quan, Z., Wang, Z.J., Ma, T., Zeng, X.: Kgnn: Knowledge graph neural network for drug-drug interaction prediction. In: IJCAI. vol.~380, pp. 2739--2745 (2020)

\bibitem{luo2021detecting}
Luo, L., Fang, Y., Cao, X., Zhang, X., Zhang, W.: Detecting communities from heterogeneous graphs: A context path-based graph neural network model. In: Proceedings of the 30th ACM international conference on information \& knowledge management. pp. 1170--1180 (2021)

\bibitem{luo2023augmented}
Luo, Z., Xu, C., Zhao, P., Geng, X., Tao, C., Ma, J., Lin, Q., Jiang, D.: Augmented large language models with parametric knowledge guiding. arXiv preprint arXiv:2305.04757  (2023)

\bibitem{mackin2021late}
Mackin, R.S., Insel, P.S., Landau, S., Bickford, D., Morin, R., Rhodes, E., Tosun, D., Rosen, H.J., Butters, M., Aisen, P., et~al.: Late-life depression is associated with reduced cortical amyloid burden: Findings from the alzheimer’s disease neuroimaging initiative depression project. Biological psychiatry  \textbf{89}(8),  757--765 (2021)

\bibitem{peng2023autorep}
Peng, H., Huang, S., Zhou, T., Luo, Y., Wang, C., Wang, Z., Zhao, J., Xie, X., Li, A., Geng, T., et~al.: Autorep: Automatic relu replacement for fast private network inference. In: Proceedings of the IEEE/CVF International Conference on Computer Vision. pp. 5178--5188 (2023)

\bibitem{peng2024maxk}
Peng, H., Xie, X., Shivdikar, K., Hasan, M.A., Zhao, J., Huang, S., Khan, O., Kaeli, D., Ding, C.: Maxk-gnn: Extremely fast gpu kernel design for accelerating graph neural networks training. In: Proceedings of the 29th ACM International Conference on Architectural Support for Programming Languages and Operating Systems, Volume 2. pp. 683--698 (2024)

\bibitem{qian2023can}
Qian, C., Tang, H., Yang, Z., Liang, H., Liu, Y.: Can large language models empower molecular property prediction? arXiv preprint arXiv:2307.07443  (2023)

\bibitem{schmidhuber2007training}
Schmidhuber, J., Wierstra, D., Gagliolo, M., Gomez, F.: Training recurrent networks by evolino. Neural computation  \textbf{19}(3),  757--779 (2007)

\bibitem{shchur2019overlapping}
Shchur, O., G{\"u}nnemann, S.: Overlapping community detection with graph neural networks. arXiv preprint arXiv:1909.12201  (2019)

\bibitem{subramaniapillai2021sex}
Subramaniapillai, S., Rajagopal, S., Snytte, J., Otto, A.R., Einstein, G., Rajah, M.N., Group, P.A.R., et~al.: Sex differences in brain aging among adults with family history of alzheimer’s disease and apoe4 genetic risk. NeuroImage: Clinical  \textbf{30},  102620 (2021)

\bibitem{tang2023signed}
Tang, H., Guo, L., Fu, X., Wang, Y., Mackin, S., Ajilore, O., Leow, A.D., Thompson, P.M., Huang, H., Zhan, L.: Signed graph representation learning for functional-to-structural brain network mapping. Medical image analysis  \textbf{83},  102674 (2023)

\bibitem{tang2022contrastive}
Tang, H., Ma, G., Guo, L., Fu, X., Huang, H., Zhan, L.: Contrastive brain network learning via hierarchical signed graph pooling model. IEEE transactions on neural networks and learning systems  (2022)

\bibitem{tang2023comprehensive}
Tang, H., Ma, G., Zhang, Y., Ye, K., Guo, L., Liu, G., Huang, Q., Wang, Y., Ajilore, O., Leow, A.D., et~al.: A comprehensive survey of complex brain network representation. Meta-Radiology p. 100046 (2023)

\bibitem{tang2021commpool}
Tang, H., Ma, G., et~al.: Commpool: An interpretable graph pooling framework for hierarchical graph representation learning. Neural Networks  \textbf{143},  669--677 (2021)

\bibitem{tzourio2002automated}
Tzourio-Mazoyer, N., Landeau, B., Papathanassiou, D., Crivello, F., Etard, O., Delcroix, N., Mazoyer, B., Joliot, M.: Automated anatomical labeling of activations in spm using a macroscopic anatomical parcellation of the mni mri single-subject brain. Neuroimage  \textbf{15}(1),  273--289 (2002)

\bibitem{vaswani2017attention}
Vaswani, A., Shazeer, N., Parmar, N., Uszkoreit, J., Jones, L., Gomez, A.N., Kaiser, {\L}., Polosukhin, I.: Attention is all you need. Advances in neural information processing systems  \textbf{30} (2017)

\bibitem{velivckovic2017graph}
Veli{\v{c}}kovi{\'c}, P., Cucurull, G., et~al.: Graph attention networks. arXiv preprint arXiv:1710.10903  (2017)

\bibitem{wang2024infuserki}
Wang, F., Bao, R., Wang, S., Yu, W., Liu, Y., Cheng, W., Chen, H.: Infuserki: Enhancing large language models with knowledge graphs via infuser-guided knowledge integration. In: Findings of the Association for Computational Linguistics: EMNLP 2024. pp. 3675--3688 (2024)

\bibitem{wang2022cell}
Wang, Y., Wang, Y.G., Hu, C., Li, M., Fan, Y., Otter, N., Sam, I., Gou, H., Hu, Y., Kwok, T., et~al.: Cell graph neural networks enable the precise prediction of patient survival in gastric cancer. NPJ precision oncology  \textbf{6}(1), ~45 (2022)

\bibitem{wang2024unlocking}
Wang, Z., Bao, R., Wu, Y., Taylor, J., Xiao, C., Zheng, F., Jiang, W., Gao, S., Zhang, Y.: Unlocking memorization in large language models with dynamic soft prompting. In: Proceedings of the 2024 Conference on Empirical Methods in Natural Language Processing. pp. 9782--9796 (2024)

\bibitem{wang2021lightweight}
Wang, Z., Wu, Y., Jia, Z., Shi, Y., Hu, J.: Lightweight run-time working memory compression for deployment of deep neural networks on resource-constrained mcus. In: Proceedings of the 26th Asia and South Pacific Design Automation Conference. pp. 607--614 (2021)

\bibitem{williamson2022sex}
Williamson, J., Yabluchanskiy, A., Mukli, P., Wu, D.H., Sonntag, W., Ciro, C., Yang, Y.: Sex differences in brain functional connectivity of hippocampus in mild cognitive impairment. Frontiers in Aging Neuroscience  \textbf{14},  959394 (2022)

\bibitem{wolterink60geometric}
WOLTERINK, J., SUK, J.: Geometric deep learning for precision medicine. KEY ENABLING TECHNOLOGY FOR SCIENTIFIC MACHINE LEARNING  \textbf{60}

\bibitem{wu2024auto}
Wu, X., Gao, S., Zhang, Z., Li, Z., Bao, R., Zhang, Y., Wang, X., Huang, H.: Auto-train-once: Controller network guided automatic network pruning from scratch. In: Proceedings of the IEEE/CVF Conference on Computer Vision and Pattern Recognition. pp. 16163--16173 (2024)

\bibitem{wu2022clare}
Wu, X., Xiong, Y., Zhang, Y., Jiao, Y., Shan, C., Sun, Y., Zhu, Y., Yu, P.S.: Clare: A semi-supervised community detection algorithm. In: Proceedings of the 28th ACM SIGKDD conference on knowledge discovery and data mining. pp. 2059--2069 (2022)

\bibitem{wu2020intermittent}
Wu, Y., Wang, Z., Jia, Z., Shi, Y., Hu, J.: Intermittent inference with nonuniformly compressed multi-exit neural network for energy harvesting powered devices. In: 2020 57th ACM/IEEE Design Automation Conference (DAC). pp.~1--6. IEEE (2020)

\bibitem{wu2020enabling}
Wu, Y., Wang, Z., Shi, Y., Hu, J.: Enabling on-device cnn training by self-supervised instance filtering and error map pruning. IEEE Transactions on Computer-Aided Design of Integrated Circuits and Systems  \textbf{39}(11),  3445--3457 (2020)

\bibitem{wu2021decentralized}
Wu, Y., Wang, Z., Zeng, D., Li, M., Shi, Y., Hu, J.: Decentralized unsupervised learning of visual representations. arXiv preprint arXiv:2111.10763  (2021)

\bibitem{wu2021enabling}
Wu, Y., Wang, Z., Zeng, D., Shi, Y., Hu, J.: Enabling on-device self-supervised contrastive learning with selective data contrast. In: 2021 58th ACM/IEEE Design Automation Conference (DAC). pp. 655--660. IEEE (2021)

\bibitem{wu2023synthetic}
Wu, Y., Wang, Z., Zeng, D., Shi, Y., Hu, J.: Synthetic data can also teach: Synthesizing effective data for unsupervised visual representation learning. In: Proceedings of the AAAI Conference on Artificial Intelligence. vol.~37, pp. 2866--2874 (2023)

\bibitem{wu2021federated}
Wu, Y., Zeng, D., Wang, Z., Sheng, Y., Yang, L., James, A.J., Shi, Y., Hu, J.: Federated contrastive learning for dermatological disease diagnosis via on-device learning. In: 2021 IEEE/ACM International Conference On Computer Aided Design (ICCAD). pp.~1--7. IEEE (2021)

\bibitem{wu2022distributed}
Wu, Y., Zeng, D., Wang, Z., Shi, Y., Hu, J.: Distributed contrastive learning for medical image segmentation. Medical Image Analysis  \textbf{81},  102564 (2022)

\bibitem{xia2021geometric}
Xia, T., Ku, W.S.: Geometric graph representation learning on protein structure prediction. In: Proceedings of the 27th ACM SIGKDD Conference on Knowledge Discovery \& Data Mining. pp. 1873--1883 (2021)

\bibitem{xie2023accel}
Xie, X., Peng, H., Hasan, A., Huang, S., Zhao, J., Fang, H., Zhang, W., Geng, T., Khan, O., Ding, C.: Accel-gcn: High-performance gpu accelerator design for graph convolution networks. In: 2023 IEEE/ACM International Conference on Computer Aided Design (ICCAD). pp. 01--09. IEEE (2023)

\bibitem{xiong2021graph}
Xiong, J., Xiong, Z., Chen, K., Jiang, H., Zheng, M.: Graph neural networks for automated de novo drug design. Drug Discovery Today  \textbf{26}(6),  1382--1393 (2021)

\bibitem{xu2018powerful}
Xu, K., Hu, W., Leskovec, J., Jegelka, S.: How powerful are graph neural networks? arXiv preprint arXiv:1810.00826  (2018)

\bibitem{ye2023natural}
Ye, R., Zhang, C., Wang, R., Xu, S., Zhang, Y.: Natural language is all a graph needs. arXiv preprint arXiv:2308.07134  (2023)

\bibitem{yu2024dynamic}
Yu, S., Bao, R., Bhatia, P., Kass-Hout, T., Zhou, J., Xiao, C.: Dynamic uncertainty ranking: Enhancing in-context learning for long-tail knowledge in llms. arXiv preprint arXiv:2410.23605  (2024)

\bibitem{zhang2023graph}
Zhang, J.: Graph-toolformer: To empower llms with graph reasoning ability via prompt augmented by chatgpt. arXiv preprint arXiv:2304.11116  (2023)

\bibitem{zhang2024pruning}
Zhang, N., Liu, Y., Zhao, X., Cheng, W., Bao, R., Zhang, R., Mitra, P., Chen, H.: Pruning as a domain-specific llm extractor. In: Findings of the Association for Computational Linguistics: NAACL 2024. pp. 1417--1428 (2024)

\bibitem{zhang2022toward}
Zhang, Y., Bao, R., Pei, J., Huang, H.: Toward unified data and algorithm fairness via adversarial data augmentation and adaptive model fine-tuning. In: 2022 IEEE International Conference on Data Mining (ICDM). pp. 1317--1322. IEEE (2022)

\bibitem{zhang2024brain}
Zhang, Y., Liu, G., Bao, R., Zhan, L., Thompson, P., Huang, H.: Brain image synthesis using incomplete multimodal data. In: 2024 IEEE International Symposium on Biomedical Imaging (ISBI). pp.~1--5. IEEE (2024)

\bibitem{zhang2024neurodegenerative}
Zhang, Y., Liu, G., Bao, R., Zhan, L., Thompson, P., Huang, H.: Neurodegenerative disease prediction via transferable deep networks. In: 2024 IEEE International Symposium on Biomedical Imaging (ISBI). pp.~1--5. IEEE (2024)

\bibitem{zhao2023survey}
Zhao, W.X., Zhou, K., Li, J., Tang, T., Wang, X., Hou, Y., Min, Y., Zhang, B., Zhang, J., Dong, Z., et~al.: A survey of large language models. arXiv preprint arXiv:2303.18223  (2023)

\end{thebibliography}

\end{document}